\documentclass{article}

\PassOptionsToPackage{numbers, compress}{natbib}

\usepackage[preprint]{neurips}

\usepackage[utf8]{inputenc} 
\usepackage[T1]{fontenc}    
\usepackage{hyperref}       
\usepackage{url}            
\usepackage{booktabs}       
\usepackage{amsfonts}       
\usepackage{nicefrac}       
\usepackage{microtype}   
\usepackage{amsmath}
\usepackage{makecell}
\usepackage{enumitem}
\usepackage{multirow}
\usepackage{xcolor}
\usepackage{graphicx}
\usepackage{pifont}
\usepackage{comment}
\newcommand{\xmark}{\ding{55}}%
\newcommand{\cmark}{\ding{51}}%
\makeatletter
\usepackage{footnote}
\usepackage[symbol]{footmisc}

\usepackage{wrapfig}

\makeatletter
\renewcommand{\paragraph}{%
  \@startsection{paragraph}{4}%
  {\z@}{0.25ex \@plus 1ex \@minus .2ex}{-1em}%
  {\normalfont\normalsize\bfseries}%
}
\makeatother

\title{ShrinkTeaNet: Million-scale Lightweight Face Recognition via Shrinking Teacher-Student Networks}

\author{
	Chi Nhan Duong\\
	Concordia University\\
	\texttt{dcnhan@ieee.org}\\
	\And
	Khoa Luu\\
	University of Arkansas\\
	\texttt{khoaluu@uark.edu}\\
	\And
	Kha Gia Quach\\
	Concordia University\\
	\texttt{kquach@ieee.org}\\
	\And
	Ngan Le\\
	Carnegie Mellon University\\
	\texttt{thihoanl@andrew.cmu.edu }
}

\begin{document}

\maketitle

\begin{abstract}
Large-scale face recognition in-the-wild has been recently achieved matured performance in many real work applications. However, such systems are built on GPU platforms and mostly deploy heavy deep network architectures. Given a high-performance heavy network as a teacher, this work presents a simple and elegant teacher-student learning paradigm, namely \textit{ShrinkTeaNet}, to train a portable student network that has significantly fewer parameters and competitive accuracy against the teacher network. Far apart from prior teacher-student frameworks mainly focusing on accuracy and compression ratios in closed-set problems, our proposed teacher-student network is proved to be more robust against open-set problem, i.e. large-scale face recognition. In addition, this work introduces a novel Angular Distillation Loss for distilling the feature direction and the sample distributions of the teacher's hypersphere to its student. Then ShrinkTeaNet framework can efficiently guide the student's learning process with the teacher's knowledge presented in both intermediate and last stages of the feature embedding. Evaluations on LFW, CFP-FP, AgeDB, IJB-B and IJB-C Janus, and MegaFace with one million distractors have demonstrated the efficiency of the proposed approach to learn robust student networks which have satisfying accuracy and compact sizes. Our ShrinkTeaNet is able to support the light-weight architecture achieving high performance with 99.77\% on LFW and 95.64\% on large-scale Megaface protocols. 
\end{abstract}

\section{Introduction}
Deep learning has been widely used in many disciplines with remarkable results comparing to previous methods. The idea of deep learning goes back to 1980s but the advent of GPUs in 2000s paved a way for using deep neural network in many areas. The breakthrough is apparent in 2012 ImageNet competition where Krizhevsky et al. presented AlexNet \cite{Alex_2012_NIPS} reducing the top-5 error with stunning result comparing the second run up. In recent time, many modern network architectures proposed that improve performance in various domains of computer vision with different applications. VGGNet \cite{simonyan2014very}, GoogLeNet \cite{Szegedy_2016_CVPR}, ResNet \cite{he2016deep} and FishNet \cite{sun2018fishnet} are prominent among other neural network architectures. Regardless of the performance improvement, these architectures conduct huge amount of computations. For example, GoogLeNet has 138 million parameters which takes few seconds to run a normal size image. Hence, such kind of networks can only be performed well using special hardware which have the capacity of parallel computing and high performance. On the other hand, such large network architectures cannot be applied on cell phones or embedded devices. Consequently, network architectures without requiring high-power machine while retaining similar accuracy are in high demand in mobile devices and embedded application.

In recent years, Knowledge Distillation has become a prominent topic by its ability to transfer the interpretation capability of a heavy network to other light architectures and make them more powerful. Given a heavy but powerful network (i.e. \textit{teacher}), conventional distillation approaches \cite{Ba_2013,Hinton_2015_NIPS} encouraged the similarity between the outputs of an input light-weight network (i.e. \textit{student}) and the teacher network to boost the student's performance. By this way, the student can efficiently inherit the power and advantages of its teacher.
Different metrics have been proposed for the similarity measurements such as $\ell_2$ or cross-entropy losses. Some other approaches injected these metrics into middle layers and exploited various aspects of the teacher network such as feature magnitude \cite{Adriana_2015_Fitnets}, feature flow \cite{Yim_2017_CVPR}, activation maps \cite{Zagoruyko_2017_AT, Heo_2018_AB}, gradient maps \cite{Guo_2018_CVPR}, and other factors \cite{Srinivas_2018_Jacobian, wang2018adversarial, Kim_2018_FT}.

Although these methods have achieved prominent results, most of them are proposed for closed-set problems where the object classes are predefined and remain the same in both training and testing phases. With this assumption, the prediction scores of the teacher can be transferred and used as supervised signals for the student. However, when the classes are different between training and testing (i.e. open-set face recognition), prediction scores of the training classes become less valuable since they are changed during testing phase. As a result, the student may have lots of difficulties when dealing with new classes. Moreover, $\ell_2$ loss are usually adopted for either the last output layer or middle layers. This is actually a hard constraint for the student while it leads toward an ``exact match'' between the features of the two networks. Therefore, when it is adopted to multiple layers, training process of the student network may have the over-regularized issues and become unstable.

\textbf{Contributions.} This paper introduces a novel teacher-student learning algorithm, namely \textit{ShrinkTeaNet}, for the open-set large-scale face recognition. The contributions of this work are three-fold.
\begin{itemize}[noitemsep,topsep=0pt]
    \item Firstly, rather than emphasizing on an exact match between features, we propose a novel Angular Distillation Loss for distilling the feature directions and the sample distributions from the teacher's hypersphere to the student. Compared to $\ell_2$ loss, the Angular constraint is ``softer'' and provides the student more flexibility to interpret the information during embedding process. Moreover, inheriting the sample distributions from its teacher can help the student robustly reuse the learned knowledge even when object classes are changed.
    \item Secondly, we present a new ShrinkTeaNet framework that efficiently distills the teacher's knowledge in every stage of the feature embedding process. 
    \item Thirdly, the evaluations show improvements in both small-scale and large-scale benchmarks.
\end{itemize}
To the best of our knowledge, this is one of the first distillation methods to tackle an open-set large-scale recognition problem.

\begin{table}[t]
	\footnotesize
	\centering
	\caption{The comparison of the properties between our distillation approach and other methods. $\ell_{CE}$ denotes the cross-entropy loss. Feature Distribution (Feat. Dist.), Activation (Act), Gradient (Grad).
	} 
	\label{tb:DistilledMethodReview} 
	\small
	\begin{tabular}{l c c c c c c}
		\Xhline{2\arrayrulewidth}
		\textbf{Method}  &
		\begin{tabular}{@{}c@{}}\textbf{Object }\\ \textbf{Class}\end{tabular}& \begin{tabular}{@{}c@{}}\textbf{Teacher}\\ \textbf{Transform}\end{tabular} & \begin{tabular}{@{}c@{}}\textbf{Student}\\ \textbf{Transform}\end{tabular}& \begin{tabular}{@{}c@{}}\textbf{Distilled}\\ \textbf{Knowledge}\end{tabular}& \begin{tabular}{@{}c@{}}\textbf{Loss}\\ \textbf{Function}\end{tabular}& \begin{tabular}{@{}c@{}}\textbf{Missing}\\ \textbf{Info.}\end{tabular}\\
		\hline
		KD \cite{Hinton_2015_NIPS} & Closed-set& Identity & Identity & Logits & $\ell_{CE}$ & \xmark\\
		FitNets \cite{Adriana_2015_Fitnets} & Closed-set& Identity & $1 \times 1$ conv & Magnitude& $\ell_2$ & \xmark\\
		Att\_Trans\cite{Zagoruyko_2017_AT} & Closed-set & Attention & Attention & Act/Grad Map& $\ell_2$ & \cmark\\
		DNN\_FSP \cite{Yim_2017_CVPR} & Closed-set & Correlation & Correlation &  Feat. Flow& $\ell_2$ & \cmark\\
		Jacob\_Match
		\cite{Srinivas_2018_Jacobian} & Closed-set & Gradient & Gradient & Jacobians& $\ell_2$ & \cmark\\
		 Factor\_Trans
		 \cite{Kim_2018_FT} & Closed-set & Encoder & Encoder & Feat. Factors& $\ell_1$ & \cmark\\
		Act\_Bound
		\cite{Heo_2018_AB} & Closed-set & Binarization & $1 \times 1$ conv & Act Map& Marginal $\ell_2$ & \cmark\\
		AL\_PSN\cite{wang2018adversarial} & Closed-set & Identity & Identity & Feat. Dist.& $\ell_{GAN}$ & \cmark\\ 
		Robust SNL\cite{Guo_2018_CVPR} & Closed-set & Identity & Identity & Scores + Grad& $\ell_2$ & \xmark\\ 
		\hline
		\hline
		\textbf{Ours} & \textbf{Open-set} & \textbf{Identity} & $\boldsymbol{1 \times 1}$ \textbf{conv} & \textbf{ Direction} & \textbf{Angular} & \xmark\\
		\Xhline{2\arrayrulewidth}
	\end{tabular}
	\vspace{-5mm}
\end{table}

\section{Related work}

There has been a rising interest in light-weight deep network design mainly in tuning deep neural architectures to strike an optimal balance between accuracy and speed. Besides directly designing new deep network architectures, depending on the motivations and techniques of the proposed frameworks, they can be divided into two different groups, i.e. network pruning and knowledge distillation.

\textbf{Network Pruning.} Network pruning is to analyze and remove the redundancy presented in the heavy network to obtain the light-weight form with comparable accuracy. Generally, the training framework for approaches in this groups consist of three stages, namely  (1) train a large (over-parameterized model), (2) prune the trained large model based on some criterions, and (3) fine-tune the pruned model to obtain good performance. There are two categories developed to prune network. The first category is  \textit{unstructured pruning methods} that starts from a large, over-parameterized and high performance model. Optimal Brain Damage \cite{Cun_1990}, Optimal brain surgeon \cite{Hassibi_1993} are first network pruning methods which are based on Hessian of the loss function. Pruning network by setting to zero the weights below a threshold, without drop of accuracy is proposed by \cite{Han_2015} and it then improved by quantizing the weights to 8 bits or less and finally Huffman encoding \cite{Han_2016}. Variational Dropout \cite{Kingma_2015} is also adopted by \cite{Molchanov_2017} to prune redundant weights. Recently, \cite{Louizos_2018} adopted sparse networks through $\ell_0$-norm regularization based on stochastic gate. In the second category, i.e. \textit{structured pruning methods}, pruning channels is a popular approach. \cite{Luo_2017} prunes filters based on its next layer by considering filter pruning as an optimization problem. The importance of a filter is calculated its absolute weights is adopted in \cite{Li_2017, Yu_2018_CVPR} to determine which channels to retain. Taylor expansion is adopted in \cite{Molchanov_2017_Talor} to approximate the importance of each channel over the final loss and prune accordingly.

\textbf{Knowledge Distillation.} Rather than trying to ``simplify'' the computationally expensive deep network as in previous group, the Knowledge Distillation approaches aim at learning a light-weight network (i.e. student) such that it can mimic the behaviors of the heavy one (i.e. teacher). With the useful information from the teacher, the student can learn more efficient and be more ``intelligent''. Inspired by this motivation, one of the first knowledge distillation works is introduced by \cite{Ba_2013} suggesting to minimize the $\ell_2$ distance between the extracted features from the last layers of these two networks. Hilton et al. \cite{Hinton_2015_NIPS} later pointed out that the hidden relationships between the predicted class probabilities from the teacher is also very important and informative for the student. Then, the soft labels generated by teacher model are adopted as the supervision signal in addition to the regular labeled training data during the training phase. In addition to the soft labels as in \cite{Hinton_2015_NIPS}, Romero et al. \cite{Adriana_2015_Fitnets} bridged the middle layers of the student and teacher networks and adopted $\ell_2$ loss to further supervised the output of the student.
Several aspects and knowledge of the teacher network are also exploited in literature including transferring the feature activation map \cite{Heo_2018_AB}, feature distribution \cite{wang2018adversarial}, block feature flow \cite{Yim_2017_CVPR}, Activation-based and Gradient-based Attention Maps \cite{Zagoruyko_2017_AT}, Jacobians \cite{Srinivas_2018_Jacobian}, Unsupervised Feature Factors \cite{Kim_2018_FT}. Recently, Guo et al. \cite{Guo_2018_CVPR} proposed to distill both prediction scores and gradient maps to enhance the student's robustness against data perturbations. Other knowledge distillation methods \cite{Yim_2017_CVPR, Tommaso_2018_ICML, Zhang_2018_CVPR, Mirzadeh_2019,Chen_2017_NIPS,Wang_2018_NIPS} are also proposed for variety of learning tasks.

\section{Proposed method}
This section firstly describes a general form of a knowledge distillation problem. Then two important design aspects for the face recognition are considered including (1) \textit{the representation of the distilled knowledge}; and (2) \textit{how to effectively transfer them between teacher and student}. Finally, the ShrinkTeaNet architecture with the Angular Distillation loss are introduced for distillation process.

Let $\mathcal{T}: \mathcal{I} \mapsto \mathcal{Z}$ and $\mathcal{S}: \mathcal{I} \mapsto \mathcal{Z}$ define the mapping functions from image domain $\mathcal{I}$ to a  high-level embedding domain. Both functions $\mathcal{T}$ and $\mathcal{S}$ are the composition of $n$ sub-functions $\mathcal{T}_i$ and $\mathcal{S}_i$ as.
\begin{equation}
\small
\begin{split}
    \mathcal{T}(I;\Theta^t) &= [\mathcal{T}_1 \circ \mathcal{T}_2 \circ \cdots \circ \mathcal{T}_n](I,\Theta^t) \\
    \mathcal{S}(I;\Theta^s) &= [\mathcal{S}_1 \circ \mathcal{S}_2 \circ \cdots \circ \mathcal{S}_n] (I,\Theta^s)
\end{split}
\end{equation}
where $I$ denotes the input image, $\Theta^t$ and $\Theta^s$ are parameters of $\mathcal{T}$ and $\mathcal{S}$, respectively. Then given a \textit{complicated high-capacity function} $\mathcal{T}$ (i.e. \textit{teacher}), the goal of model distillation is to distill the knowledge from $\mathcal{T}$ to a \textit{limited-capacity function} $\mathcal{S}$ (i.e. \textit{student}) so that $\mathcal{S}$ can embed similar latent domain as $\mathcal{T}$.
In order to achieve this goal, the learning process of $\mathcal{S}$ is usually taken place under the supervision of $\mathcal{T}$ by comparing the their outputs step-by-step. 
\begin{equation} \label{eqn:LossDistill}
\small
\begin{split}
    \mathcal{L}_i(\mathcal{S},\mathcal{T}) &= d \left(\mathcal{G}^t_i(F^t_i), \mathcal{G}^s_i(F^s_i) \right), i=1 .. n \\ 
    F^t_i &= \left[\mathcal{T}_1 \circ \mathcal{T}_2 \circ \cdots \circ \mathcal{T}_i\right](I,\Theta^t)\\
    F^s_i &= \left[\mathcal{S}_1 \circ \mathcal{S}_2 \circ \cdots \circ \mathcal{S}_i\right](I,\Theta^s) 
\end{split}
\end{equation}
where $\mathcal{G}^t_i(\cdot)$ and $\mathcal{G}^s_i(\cdot)$ are transformation functions of $\mathcal{T}$ and $\mathcal{S}$ making their corresponding embedded features comparable. $d(\cdot,\cdot)$ denotes the difference between these transformed features.
Then by minimizing these differences $\mathcal{L}_{distill} = \sum_i^n \lambda_i \mathcal{L}_i(\mathcal{S},\mathcal{T})$, the knowledge from the teacher $\mathcal{T}$ can be transferred to the student $\mathcal{S}$ so that they can embed similar latent domain.
It is worth noting that the form of $\mathcal{L}_i(\mathcal{S},\mathcal{T})$ 
provides two important properties. Firstly, since $d(\cdot,\cdot)$ measures the distance between $F_i^t$ and $F_i^s$, it implicitly defines the knowledge to be transferred from $\mathcal{T}$ to $\mathcal{S}$. Secondly, the transformation functions $\mathcal{G}^t_i(\cdot)$ and $\mathcal{G}^s_i(\cdot)$ controls the portion of the transferred information.
The next sections focus on the designs of these two components for selecting the most useful information and transferring them to the student without missing important information from the teacher. 

\subsection{Distilled Knowledge from Teacher Hypersphere}
As presented in Table \ref{tb:DistilledMethodReview}, most previous distillation frameworks are introduced for the closed set classification problem, i.e. object classification or semantic segmentation with predefined classes. With the assumption about the fixed (and small) number of classes, traditional metrics can be efficiently adopted for distillation process. 
For example, the $\ell_2$ distance can be used for similarity measurement between $\mathcal{S}$ and $\mathcal{T}$, i.e. $d(\mathcal{G}^t_i(F^t_i), \mathcal{G}^s_i(F^s_i)) = \parallel \mathcal{G}^t_i(F^t_i) -  \mathcal{G}^s_i(F^s_i) \parallel^2_2$. However, since the capacity of $\mathcal{S}$ is limited, employing this constraint as a regularization to each $F^s_i$ (i.e. to enforce $F^s_i$ and $F^t_i$ to be exact matched)
can lead to the over-regularized issue.Consequently, this constraint becomes too hard and makes the learning process of $\mathcal{S}$ more difficult.
Another metric is to adopt the class probabilities predicted from the teacher $\mathcal{T}$ as the soft target distribution for the student $\mathcal{S}$ \cite{Hinton_2015_NIPS}. However, this metric is efficient only when the object classes are fixed in both training and testing phases. Otherwise, the knowledge to convert embedded features to class probabilities cannot be reused during testing stage and, therefore, the distilled knowledge is also partially ignored.

In open-set problems, since classes are not predefined beforehand, the sample distributions of each class and the margin between classes become more valuable knowledge. In other words, with the open-set problems, the angular differences between samples and how the samples distributed in the teacher's hypersphere are more beneficial for the student.
Therefore, we propose to use the angular information as the main knowledge to be distilled. 
By this way, rather than enforcing the student follow the exact outputs of the teacher (as in case of $\ell_2$ distance), we can relax the constraint so that the embedded features extracted by the student only need to have similar direction as those extracted by the teacher.
Generally, with ``softer'' the distillation constraint, the student is able to adaptively interpret the teacher's information and learn the solution process more efficiently.

\begin{figure}[t]
	\centering \includegraphics[width=0.85\columnwidth]{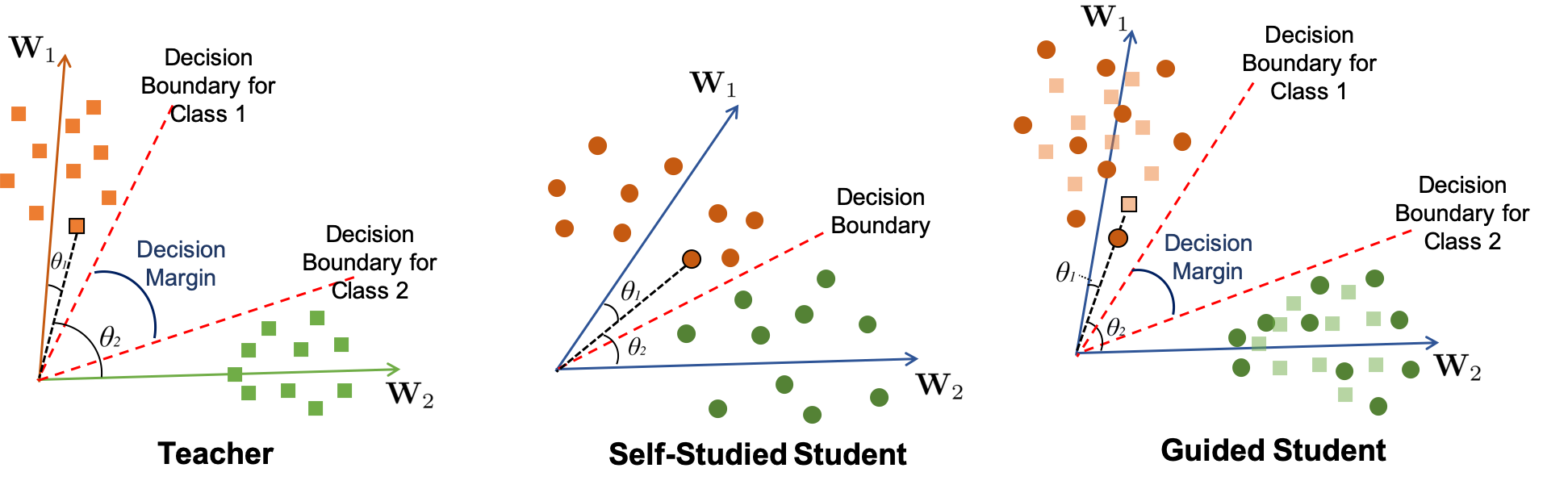}
	\caption{\textbf{Geometric Interpretation of Angular Distillation Loss.} With high-capacity function $\mathcal{T}$, the teacher are able to produce a large decision margin between two classes while the self-studied student only gives small decision margin. By following the direction provided by the teacher, the student can make better decision boundaries with larger margin between classes. 
	Moreover, with the angular distillation loss, the student is not strictly required to produce exact feature as its teacher (\textit{i.e. which is a very hard constraint according to the student's capability)}.
	}
	\label{fig:GeometricInterpretation}
	\vspace{-5mm}
\end{figure}

\textbf{Softmax Loss Revisit.} As one of the most widely used losses for classification problem, Softmax loss of each input image is formulated as follows. 
\begin{equation}
\small
    \mathcal{L}_{SM} = -\log \frac{e^{\mathbf{W}_y* F^s_n}}{\sum_{c=1}^C e^{\mathbf{W}_c * F^s_n} } = -\log \frac{e^{\parallel\mathbf{W}_y\parallel \parallel F^s_n \parallel \cos \theta_y}}{\sum_{c=1}^C e^{\parallel \mathbf{W}_c\parallel \parallel F^s_n\parallel \cos \theta_c} } \label{eqn:SMLoss}
\end{equation}
where $y$ is the index of the correct class of the input image and $C$ denotes the number of classes. Notice that the bias term is fixed to $0$ for simplicity.
By adopting the $\ell_2$ normalisation to both feature $F^s_n$ and weight $\mathbf{W}_c$, the angle between them becomes the only classification criteria. If each weight vector $\mathbf{W}_c$ is regarded as the representative of class $c$, minimizing the loss means that the samples of each class are required to distributed around that class' representative with the minimal angular difference.
This is also true during testing process where the angle between the direction of extracted features from input image and the representative of each class (i.e. classification problem with predefined classes) or  extracted features of other sample (i.e. verification problem) are used for deciding whether they belong to the same class. 
In this respect, the magnitude of the feature $F^s_n$ becomes less important than its direction. Therefore, rather than considering both magnitude and direction of $F_n^t$ for distillation process, knowledge about the direction is enough for the student to achieve similar distribution as the teacher's hypersphere. Moreover, this knowledge can be also efficiently reused to compare samples of object classes other than the ones in training.

\textbf{Feature Direction as Distilled Knowledge.} 
We propose to use the direction of the teacher feature $F^t_n$ as the distilled knowledge and define an \textbf{angular distillation loss} as follows.
\begin{equation}
\small
    \mathcal{L}_n(\mathcal{S}, \mathcal{T}) = d(\mathcal{G}^t_n(F^t_n), \mathcal{G}^s_n(F^s_n)) = \left\| 1 - \frac{\mathcal{G}^t_n(F^t_n)}{\parallel \mathcal{G}^t_n(F^t_n)\parallel}* \frac{\mathcal{G}^s_n(F^s_n)}{\parallel \mathcal{G}^s_n(F^s_n)\parallel}\right\| ^2_2
\end{equation}
With this form of distillation, the only knowledge needed to be transferred between $\mathcal{T}$ and $\mathcal{S}$ is the direction of embedded features. In other words, as long as $F^s_n$ and $F^t_n$ have similar direction, these features can freely distributed on different hyperspheres with various radius in latent space. This produces a degree of freedom for $\mathcal{S}$ to interpret its teacher's knowledge during learning process. 
Incorporating this distillation loss to Eqn. (\ref{eqn:SMLoss}), the objective function becomes.
\begin{equation} \label{eqn:TotalLoss}
    \mathcal{L} = \mathcal{L}_{SM} + \lambda_n \mathcal{L}_n(\mathcal{S}, \mathcal{T})
\end{equation}
The first term corresponds to the traditional classification loss whereas the second term guides the student to learn from the hypersphere of the teacher. Notice that, this objective function is not limited to specific classification loss. This distillation loss can act as a support to any other loss functions.

\textbf{The Transformation Functions.} To prevent missing information during distillation, we choose identity transformation for $\mathcal{T}$, i.e. $\mathcal{G}^t_n(F^t_n) = F^t_n$, while  $\mathcal{G}^s_n(F^s_n)$ is chosen as a mapping function $\mathcal{H}^s_n: \mathbb{R}^{h_n \times w_n \times m^s_n} \mapsto \mathbb{R}^{h_n \times w_n \times m^t_n}$ such that the dimension of $F^s_n$ is increased to match the dimension of $F^t_n$. 
For example, if 
$F^s_n \in \mathbb{R}^{h_n \times w_n \times m^s_n} $
is a feature map extracted from a Deep Neural Network, $\mathcal{H}^s_n(F^s_n)$ can be defined as an $1 \times 1$ convolution layer to transform $F^s_n$ to $\bar{F}^s_n \in \mathbb{R}^{h_n \times w_n \times m^t_n}$ where $h_n \times w_n \times m^t_n$ is the dimension of $F^t_n$. By this way, no information is missing during feature transformation and, therefore, $\mathcal{S}$ can take full advantages of all knowledge from $\mathcal{T}$.

\begin{figure}[t]
	\centering \includegraphics[width=0.85\columnwidth]{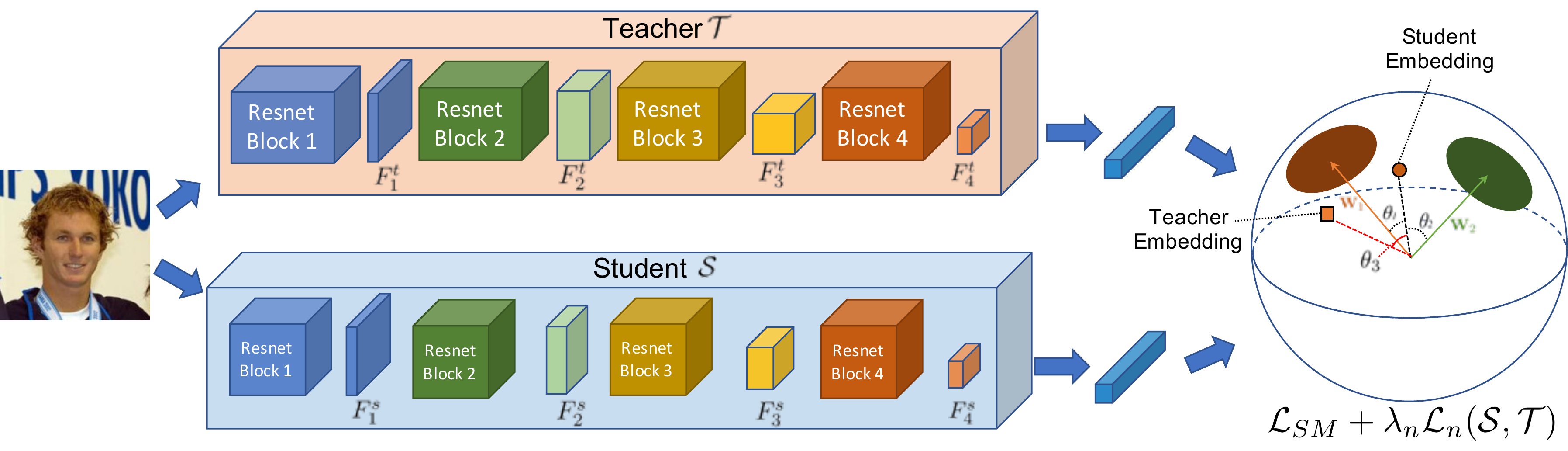}
	\caption{\textbf{ShrinkTeaNet Architecture For Knowledge Distillation in Last Layer.} Given an input image, the feature directions provided by the student is optimized using both directions of the class' representative and the teacher embedding.}
	\label{fig:ShrinkTeaNetLastLayer}
	\vspace{-5mm}
\end{figure}

\textbf{Geometric Interpretation.}
Considering the binary classification where there are only two classes with the representatives $\mathbf{W}_1$ and $\mathbf{W}_2$.
As presented in previous section, when both embedded features and the representatives are normalized, 
the classification results depend entirely on the angle between them, i.e. $\theta_1$ and $\theta_2$.
In the training stage, the softmax loss $\mathcal{L}_{SM}$ requires $\theta_1 < \theta_2$ to classify an input image as class $1$ and vice versa. 
As illustrated in Figure \ref{fig:GeometricInterpretation}, with higher capacity, the teacher can provide better decision margin between the two classes, while the self-studied student (i.e. using only softmax loss function) can only give small decision margin.
When the $\mathcal{L}_n(\mathcal{S}, \mathcal{T})$ is incorporated (i.e. guided student), the classification margin between class 1 and class 2 is further enhanced by following the feature directions of its teacher and produces better decision boundaries. 
Furthermore, one can easily see that even when the student is not able to produce exact features as its teacher, it can easily mimic the teacher's feature directions and be beneficial from the teacher's hypersphere.

\subsection{Intermediate Distilled Knowledge}
In this section, we further distill the knowledge of the teacher to intermediate components of the student. 
Generally, if the input to $\mathcal{S}$ is interpreted as the question and the distribution of its embedded features is its answer, the generated features at the middle stage, i.e. $F^s_i$, can be viewed as the 
intermediate understanding or interpretation of the student about the solution process. Then to help the student efficiently ``understand'' the development of the solution, the teacher should illustrate to the student \textit{``how the good features look like''} and \textit{``whether the current features of the student is good enough to get the solution in later steps''}. Then, the teacher can supervise and efficiently correct the student from the beginning and, therefore, leading to more efficient learning process of the student.

Similar to previous section, rather than employing $\ell_2$ norm as the cost function of each pair $\{F^s_i,F^t_i\}$, we proposed to validate the quality of $F^s_i$ based on its angular difference between the embedding produced by $F^s_i$ and $F^t_i$ using the same teacher's interpretation toward the last stages.
In particular, the distillation loss for each intermediate feature $F^s_i$ can be formulated as follows.
\begin{equation} \label{eqn:DistillIntermediate}
\small
\begin{split}
    \mathcal{L}_i(\mathcal{S}, \mathcal{T}) = & d(\mathcal{G}^t_i(F^t_i), \mathcal{G}^s_i(F^s_i))\\
    = & d(\left[\mathcal{T}_{i+1}\circ \cdots \circ \mathcal{T}_n \right](F^t_i), \left[\mathcal{H}^s_i \circ \mathcal{T}_{i+1}\circ \cdots \circ \mathcal{T}_n \right](F^s_i))\\
    = & \left\| 1 - \frac{\left[\mathcal{T}_{i+1}\circ \cdots \circ \mathcal{T}_n \right](F^t_i)}{\parallel \left[\mathcal{T}_{i+1}\circ \cdots \circ \mathcal{T}_n \right](F^t_i)\parallel}* \frac{\left[\mathcal{H}^s_i \circ \mathcal{T}_{i+1}\circ \cdots \circ \mathcal{T}_n \right](F^s_i)}{\parallel \left[\mathcal{H}^s_i \circ \mathcal{T}_{i+1}\circ \cdots \circ \mathcal{T}_n \right](F^s_i)\parallel}\right\| ^2_2 
\end{split}
\end{equation}
The intuition behind this distillation loss for each intermediate feature $F^s_i$ is to validate whether $F^s_i$ contains enough useful information to make the similar decision as its teacher in the later steps.
In order to validate this point, we propose to take advantage of the teacher power to solve the solution given the student input at $i$-th stage, i.e. $F^s_i$. In case the teacher can still get similar solution using that input, then the student's understanding until that stage is acceptable. Otherwise, the student is required to be re-corrected immediately. 
As a result, given the intermediate feature $F^s_i$, it is firstly transformed by $\mathcal{H}^s_i$ to match the dimension of the teacher feature $F^t_i$. Then both $F^s_i$ and $F^t_i$ are analyzed by the teacher $\mathcal{T}$, i.e. $\left[\mathcal{T}_{i+1}\circ \cdots \circ \mathcal{T}_n \right]$, for the final embedding features. Finally, their similarity in the hypersphere is used for validate the knowledge that $F^s_i$ embeds.

\begin{figure}[t]
	\centering \includegraphics[width=0.9\columnwidth]{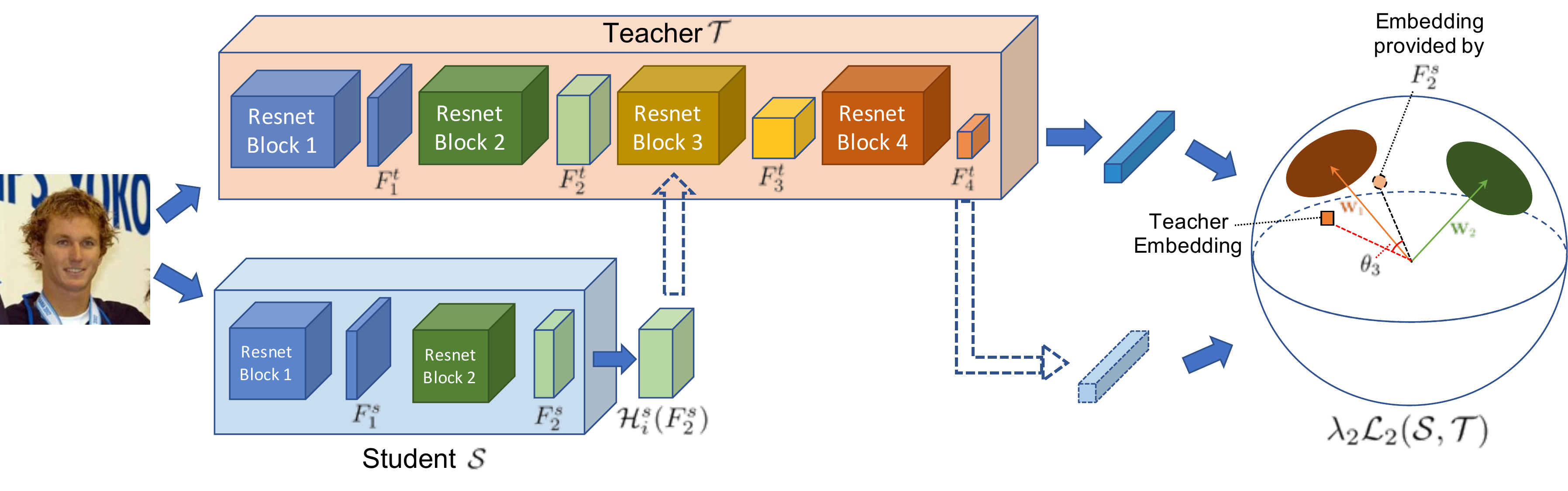}
	\caption{\textbf{ShrinkTeaNet Architecture for Intermediate Knowledge Distillation.} Given an intermediate student's features, i.e. $\mathcal{H}_i^s(F_i^s)$, the teacher use its power to validate whether the student's input is informative enough to make similar decision as the teacher.}
	\label{fig:ShrinkTeaNetIntermedidate}
	\vspace{-5mm}
\end{figure}

\subsection{Shrinking Teacher-Student Network for Face Recognition}

Figures \ref{fig:ShrinkTeaNetLastLayer} and \ref{fig:ShrinkTeaNetIntermedidate} illustrate our proposed ShrinkTeaNet framework to distill the knowledge for both final and intermediate features. Resnet-style convolution neural networks (CNN) with four resnet blocks are adopted for both teacher and student. The whole networks can be considered as the mapping functions, i.e. $\mathcal{T}$ and $\mathcal{S}$, whereas each resnet block corresponds to each sub-function, i.e. $\mathcal{T}_i$ and $\mathcal{S}_i$.
Then we learn strong but efficient student network by distilling the knowledge to all four blocks.
Given the dataset $D=\{I_j,y_j\}_{j=1}^N$ consisting of $N$ facial images $I_j$ and their corresponding labels $y_j$.
The overall learning objective can be formulated as the composition of Eqns. (\ref{eqn:TotalLoss}) and (\ref{eqn:DistillIntermediate}).
\begin{equation} \label{eqn:OverallLoss}
\small
    \mathcal{L} = \frac{1}{N} \sum_j \left[ \mathcal{L}_{SM} + \sum_{i=1}^n \lambda_i \mathcal{L}_i(\mathcal{S}, \mathcal{T})\right]
\end{equation}
where $\lambda_i$ denotes the hyper-parameter control the balance between the distilled knowledge to be transferred at different resnet blocks. Moreover, as presented in previous section, the identity transformation function is used for all functions $G^t_i$ while $1 \times 1$ convolution followed by batch normalization layer is adopted for $G^s_i$ to match the dimension of the corresponding teacher's features.
\begin{table}[t]
	\footnotesize
	\centering
	\caption{Verification performance (\%) on different small-scale datasets, i.e. LFW, CFP-FP, AgeDB. 
	} 
	\label{tb:SmallScaleBenchmark} 
	\footnotesize
	\begin{tabular}{c c c l c c c }
		\Xhline{2\arrayrulewidth}
		\textbf{Backbone} & \begin{tabular}{@{}c@{}}\textbf{\# of} \\ \textbf{params}\end{tabular} & \textbf{Ratio} & \textbf{Model Type} &
		\begin{tabular}{@{}c@{}}\textbf{LFW}\end{tabular}& \begin{tabular}{@{}c@{}}\textbf{CFP-FP}\\ \end{tabular} & \begin{tabular}{@{}c@{}}\textbf{AgeDB}\\ \end{tabular}\\
		\hline
		\begin{tabular}{@{}l@{}} ResNet90 \cite{he2016deep}\end{tabular} &  63.67M & 100\% &Teacher & 99.82\% & 96.83\% & 98.37\%\\
		\hline
		\multirow{3}{*}{\begin{tabular}{@{}c@{}}MobileNetV1 \cite{howard2017mobilenets} \\ (MV1)\end{tabular}} 
		& \multirow{3}{*}{3.53M}& \multirow{3}{*}{5.54\%} & Self-Studied  & 99.53\% & 93.81\% & 96.30\%\\
		& & & Student-1 ($\ell_2$ loss)&  99.60\%& 93.39\% & 96.83\%\\
		& & & \textbf{ShrinkTeaNet-MV1}& \textbf{99.63\%} & \textbf{94.23\%} & \textbf{97.10\%}\\
		\hline
		\multirow{3}{*}{\begin{tabular}{@{}c@{}}MobileNetV2 \cite{sandler2018mobilenetv2} \\ (MV2)\end{tabular}} & \multirow{3}{*}{2.15M} &  \multirow{3}{*}{3.38\%} & Self-Studied  &  99.42\%& 91.67\% & 95.28\%\\
		& & & Student-1 ($\ell_2$ loss)& 99.53\% & 91.03\% & 96.20\%\\
		& & & \textbf{ShrinkTeaNet-MV2}& \textbf{99.63\%} & \textbf{93.29\%} & \textbf{97.00\%}\\
		 \hline
		 \multirow{3}{*}{\begin{tabular}{@{}c@{}}MobileFacenet \cite{chen2018mobilefacenet} \\ (MFN)\end{tabular}} & \multirow{3}{*}{1.2M} & \multirow{3}{*}{1.88\%} &Self-Studied & 99.45\% & 92.11\% & 96.17\%\\
		 & & & Student-1 ($\ell_2$ loss)& 99.50\% & 91.93\% & 96.45\%\\
		 & & & \textbf{ShrinkTeaNet-MFN} & \textbf{99.60\%} & \textbf{93.44\%} & \textbf{96.73\%}\\
		 \hline
		 \multirow{3}{*}{\begin{tabular}{@{}c@{}}MobileFacenet-R \\ (MFNR)\end{tabular}} & \multirow{3}{*}{3.73M} & \multirow{3}{*}{5.86\%}& Self-Studied & 99.60\% & 93.80\% & 96.90\%\\
		 & & & Student-1 ($\ell_2$ loss)& 99.68\% &  94.51\% & 97.48\%\\
		 & & & \textbf{ShrinkTeaNet-MFNR} & \textbf{99.77\%} & \textbf{95.14\%} & \textbf{97.63\%}\\
		\Xhline{2\arrayrulewidth}
	\end{tabular}
	\vspace{-6mm}
\end{table}
\section{Experimental results}
\subsection{Databases}
\begin{wraptable}{r}{0.48\textwidth}
	\caption{Verification performance (\%) on LFW.}
	\label{tb:LFWBenchmark} 
	\small
	\begin{tabular}{l c c}
		\Xhline{2\arrayrulewidth}
		\textbf{Method}  &
		\begin{tabular}{@{}c@{}}\textbf{Training}\\\textbf{Data}\end{tabular}& \begin{tabular}{@{}c@{}}\textbf{Accuracy}\\ \end{tabular} \\
		\hline
		Center Loss \cite{wen2016discriminative} & 0.7M&  99.28\%\\
		Sphereface \cite{liu2017sphereface} & 0.5M & 99.42\% \\
		Sphereface+ \cite{liu2018learning} & 0.5M & 99.47\% \\
		Deep-ID2+ \cite{sun2014deep} & 0.3M & 99.47\%\\
		Marginal Loss \cite{Wang_2018_CVPR}& 4M & 99.48\%\\
		RangeLoss \cite{zhang2017range} & 5M & 99.52\%\\
		FaceNet \cite{schroff2015facenet} & 200M & 99.63\%\\
		CosFace \cite{Wang_2018_CVPR}& 5M & 99.73\%\\
		ArcFace \cite{deng2018arcface} & 5.8M & 99.83\% \\
		\hline \hline
		ShrinkTeaNet-MV1 & 5.8M & 99.63\%\\
		ShrinkTeaNet-MV2 & 5.8M & 99.63\%\\
		ShrinkTeaNet-MFN & 5.8M & 99.60\%\\
		\textbf{ShrinkTeaNet-MFNR} & 5.8M & \textbf{99.77\%}\\
		\Xhline{2\arrayrulewidth}
	\end{tabular}
	\vspace{-10mm}
\end{wraptable}

\textbf{MS-Celeb-1M} \cite{guo2016ms} consists of 10M photos of 100K subjects. However, a large portion of this dataset includes noisy images or incorrect ID labels. A cleaned version of this dataset \cite{deng2018arcface} is provided with 5.8M photos from 85K identities.

\textbf{Labeled Faces in the Wild (LFW)} \cite{huang2008labeled} is introduced with 13,233 in-the-wild facial images of 5749 subjects. They are divided into 6000 matching pairs with 3000 positive matches.

\textbf{MegaFace} \cite{kemelmacher2016megaface} provides a very challenging testing protocol with million-scale of distractors. The gallery set includes more than 1 million images of 690K subjects while the probe set consists of 100K photos from 530 identities.

\textbf{Celebrities Frontal-Profile} \cite{sengupta2016frontal} was released to validate the models on frontal vs profile modes. It consists of 7000 matching pairs from 500 subjects. 

\textbf{AgeDB} \cite{moschoglou2017agedb} 
provides a protocol with 4 testing age groups where each group consists of 10 splits of 600 matching pairs from 440 subjects. Besides age factor, other facial variations (i.e. pose, illumination, expression) are also included.

\textbf{IJB-B} \cite{whitelam2017iarpa} and \textbf{IJB-C} \cite{maze2018iarpa} are  introduced as two large-scale face verification protocols. While IJC-B provides 12115 templates with 10270 positive matches and 8M negative matches, its extension, i.e. IJB-C, further provides 23124 templates with 19557 positive and 15639K negative matching pairs.

\subsection{Implementation Details}
\textbf{Data Preprocessing.} All faces are firsly detected using MTCNN \cite{MTCNN} and aligned to a predefined template using similarity transformation. They are then cropped to the size of $112 \times 112$.

\textbf{Network Architectures.}
For all the experiments, we use the Resnet-90 structure \cite{he2016deep} as the teacher network, while different light-weight networks, i.e. MobileNetV1 \cite{howard2017mobilenets}, MobileNetV2 \cite{sandler2018mobilenetv2}, MobileFaceNet \cite{chen2018mobilefacenet}. A modified version of MobileFacenet, namely MobileFacenet-R, is also adopted for the student network. This modified version is similar to MobileFacenet except the feature size of each resnet-block is equal to the size of its corresponding features in the teacher network. 

\textbf{Model Configurations.} 
In training stage, the batch size is set to 512. The learning rate starts from 0.1 and the momentum is 0.9. All the models are trained in MXNET environment with a machine of Core i7-6850K @3.6GHz CPU, 64.00 GB RAM with four P6000 GPUs.
The $\lambda_n$ is experimentally set to 1 in case of Angular Distillation Loss while this parameter is set to 0.001 as the case of $\ell_2$ loss due to the large value of the loss with large feature map. For the intermediate layers, $\lambda_i = \frac{\lambda_{i+1}}{2}$.

\subsection{Evaluation Results}
\textbf{Small-scale Protocols.} We validate the efficiency of our ShrinkTeaNet framework with four light-weight backbones on small scale protocols. The Resnet-90 trained on MS-Celeb-1M acts as the teacher network. Then, for each light-weight backbone, three cases are considered: (1) Self-studied student which is trained without the help from the teacher; (2) Student-1 which is trained using the objective function as Eqn. (\ref{eqn:OverallLoss}) but the $\ell_2$ function is adopted for distillation loss; and (3) our ShrinkTeaNet with Angular Distillation loss.
Table \ref{tb:SmallScaleBenchmark} illustrates the performance of the teacher network together with its students. Due to the limited-capacity of the light-weight backbones, in all four cases, the self-studied networks leave the performance gaps of 0.2\% $-$ 0.4\%, 3.02\% $-$ 5.16\%, and 1.47\% $-$ 3.09\% with their teacher on LFW, CFP-FP, and AgeDB, respectively. 
Although the guided students using $\ell_2$ loss can slightly improve the accuracy, it is not always the case when the accuracy of MobileNetV1, MobileNetV2, and MobileFacenet are reduced in CFP-FP benchmark. Moreover, we also notice that the training process with $\ell_2$ loss is unstable.
Meanwhile, our proposed ShrinkTeaNet efficiently distills the knowledge from the teacher network to its student with
the best performance gaps that are significantly reduced to only 0.05\%, 1.83\%, and 0.74\% on the three benchmarks LFW, CFP-FP, AgeDB, respectively. The comparisons with other face recognition methods against LFW are presented in Table \ref{tb:LFWBenchmark}. From these results, even with the light-weight backbone, ShrinkTeaNet can achieve competitive performance with other large-scale networks.

\begin{savenotes}
\begin{table}[t]
	\footnotesize
	\begin{minipage}{0.48\textwidth}
	\centering
	\caption{Comparison with different methods on Megaface Challenge 1 protocol. 
	} 
	\label{tb:MegafaceBenchmark} 
	\footnotesize
	\begin{tabular}{ >{\arraybackslash}m{3cm} c c}
		\Xhline{2\arrayrulewidth}
		\textbf{Method}  &
		\begin{tabular}{@{}c@{}}\textbf{Protocol}\end{tabular}& \begin{tabular}{@{}l@{}}\textbf{Accuracy}\\ \end{tabular} \\
		\hline
		Sphereface \cite{liu2017sphereface} & Small & 72.73\% \space \space \\
		Sphereface+ \cite{liu2018learning} & Small & 73.03\% \space \space \\
		Center Loss \cite{wen2016discriminative} & Small&  65.49\% \space \space\\
		\hline
		FaceNet \cite{schroff2015facenet} & Large & 70.49\% \space \space\\
		CosFace \cite{Wang_2018_CVPR}& Large & 82.72\% \space \space\\
		ArcFace \cite{deng2018arcface} & Large & 98.35\%\footnote{ refers to the accuracy obtained by using the refined testing dataset with cleaned labels from \cite{deng2018arcface}.}\\
		\hline \hline
		MV1 \cite{howard2017mobilenets} & Large & 91.93\%$^*$\\
		\textbf{ShrinkTeaNet-MV1} & Large & \textbf{94.16\%}$^*$\\
		\hline
		MV2 \cite{sandler2018mobilenetv2} & Large & 89.22\%$^*$\\
		\textbf{ShrinkTeaNet-MV2} & Large & \textbf{92.86\%}$^*$\\
		\hline
		MFN \cite{chen2018mobilefacenet} & Large & 89.32\%$^*$\\
		\textbf{ShrinkTeaNet-MFN} & Large & \textbf{91.89\%}$^*$\\
		\hline
		MFNR & Large & 93.74\%$^*$\\
		\textbf{ShrinkTeaNet-MFNR} & Large &\textbf{ 95.64\%}$^*$\\
		\Xhline{2\arrayrulewidth}
	\end{tabular}
	\end{minipage} 
	\hspace{0.15cm}
	\begin{minipage}{0.5\textwidth}
	\centering
   	\caption{Comparison with different methods on 1:1 IJB-B and IJB-C Verification protocol. The accuracy is reported at TAR (@FAR=1e-4).} 
	\label{tb:IJBBenchmark} 
	\footnotesize
	\begin{tabular}{l c c c}
		\Xhline{2\arrayrulewidth}
		\textbf{Method}  & \begin{tabular}{@{}c@{}}\textbf{Training} \\ \textbf{Data}\end{tabular} &
		\begin{tabular}{@{}c@{}}\textbf{IJB-B}\end{tabular}& \begin{tabular}{@{}c@{}}\textbf{IJB-C}\\ \end{tabular} \\
		\hline
		SENet50 \cite{cao2018vggface2} & VGG2 & 0.800 & 0.840 \\
		MN-VC \cite{xie2018multicolumn} & VGG2 & 0.831 & 0.862\\
		ResNet50 + DCN \cite{Xie18a} & VGG2 & 0.841 & 0.880\\
		ArcFace \cite{deng2018arcface} & VGG2& 0.898 & 0.921\\
		ArcFace \cite{deng2018arcface} & MS1M & 0.942 & 0.956\\
		\hline \hline
		MV1 \cite{howard2017mobilenets} & MS1M & 0.909 &0.930\\
		\textbf{ShrinkTeaNet-MV1} & MS1M & \textbf{0.915} &\textbf{0.936}\\
		\hline
		MV2 \cite{sandler2018mobilenetv2} & MS1M & 0.883 &0.907\\
		\textbf{ShrinkTeaNet-MV2} & MS1M & \textbf{0.902} &\textbf{0.922}\\
		\hline
		MFN \cite{chen2018mobilefacenet} & MS1M & 0.896&0.918\\
		\textbf{ShrinkTeaNet-MFN} & MS1M & \textbf{0.898} &\textbf{0.921}\\
		\hline
		MFNR & MS1M & 0.912& 0.916\\
		\textbf{ShrinkTeaNet-MFNR} & MS1M & \textbf{0.923} &\textbf{0.940}\\
		\Xhline{2\arrayrulewidth}
	\end{tabular}
	\end{minipage}
	\vspace{-5mm}
\end{table}
\end{savenotes}
\textbf{Megaface Protocol.} We adopt similar training process as in small-scale protocols and evaluate our ShrinkTeaNet on the challenging Megaface benchmark against millions of distractors. The comparison in terms of the Rank-1 identification rates between our ShrinkTeaNet and other models is presented in Table \ref{tb:MegafaceBenchmark}. These results again show the advantages of our ShrinkTeaNet with consistent improvements provided to the four light-weight student networks. The performance gains for the MV1, MV2, MFN, and MFNR are 2.23\%, 3.64\%, 2.57\%, and 1.9\%, respectively. Moreover, the ShrinkTeaNet-MFNR achieves competitive accuracy (i.e. 95.64\%) with other large-scale network and reduces the gap with ArcFace \cite{deng2018arcface} to only 1.71\%.

\textbf{IJB-B and IJB-C Protocols.} The comparisons against other recent methods on IJB-B and IJB-C benchmarks are also illustrated in Table \ref{tb:IJBBenchmark}. Similar to the Megaface protocols, ShrinkTeaNet is able to boost the performance of the light-weight backbones significantly and reduce the performance gap to the large-scale backbone to only 0.019 on IJB-B and 0.016 on IJB-C. 
These results have further emphasized the advantages of the proposed ShrinkTeaNet framework for model distillation.

\large
\paragraph{Conclusions.}
\normalsize
This paper presents a novel teacher-student learning paradigm, namely ShrinkTeaNet, for open-set face recognition. By adopting the proposed Angular Distillation Loss with the distillation on every stage of feature embbeding process, the student network can absorb the knowledge of the teacher's hypersphere in an relaxing and efficient manners. These learned knowledge can flexible adopt even when the testing classes are different from the training ones. Evaluation in both small-scale and large-scale protocols showed the advantages of the proposed ShrinkTeaNet framework.

\small

\bibliographystyle{unsrt}
\bibliography{ref}

\end{document}